\documentclass[10pt,twocolumn,letterpaper]{article}

\usepackage{cvpr}              % To produce the CAMERA-READY version
\usepackage{graphicx}
\usepackage{amsmath}
\usepackage{amssymb}
\usepackage{booktabs,caption}
\usepackage{multirow}
\usepackage{url}
\usepackage{bm}
\usepackage{dsfont}
\usepackage[linesnumbered,ruled,vlined]{algorithm2e}
\usepackage{makecell}
\usepackage[flushleft]{threeparttable}
\usepackage{enumitem}
\usepackage[accsupp]{axessibility} % Improves PDF readability for those with disabilities.
\usepackage{siunitx}
\usepackage[dvipsnames]{xcolor}
\usepackage[table]{xcolor}

\graphicspath{{figures/}}
\definecolor{iccvblue}{rgb}{0.21,0.49,0.74}
\usepackage[pagebackref,breaklinks,colorlinks,allcolors=iccvblue]{hyperref}

\usepackage[capitalize]{cleveref}
\crefname{section}{Sec.}{Secs.}
\Crefname{section}{Section}{Sections}
\Crefname{table}{Table}{Tables}
\crefname{table}{Tab.}{Tabs.}

\usepackage{indentfirst}

\def\paperID{22}
\def\confName{ICCV}
\def\confYear{2025}

\begin{document}

\title{AIM 2025 Challenge on High FPS Motion Deblurring: Methods and Results}

\author{
George Ciubotariu$^\dagger$ \and
Florin-Alexandru Vasluianu$^\dagger$ \and
Zhuyun Zhou$^\dagger$ \and
Nancy Mehta$^\dagger$ \and
Radu Timofte$^\dagger$ \and
% VPEG
Ke Wu \and
Long Sun \and
Lingshun Kong \and
Zhongbao Yang \and
Jinshan Pan \and
Jiangxin Dong \and
Jinhui Tang \and
% VPEG_2_3
Hao Chen \and
Yinghui Fang \and
% SRC-B
Dafeng Zhang \and
Yongqi Song \and
Jiangbo Guo \and
Shuhua Jin \and
% X-L
Zeyu Xiao \and
Rui Zhao \and
Zhuoyuan Li \and
Cong Zhang \and
% BlurKing_Sharper
Yufeng Peng \and
Xin Lu \and
Zhijing Sun \and
Chengjie Ge \and
Zihao Li \and
Zishun Liao \and
Ziang Zhou \and
Qiyu Kang \and
Xueyang Fu \and
Zheng-Jun Zha \and
% Mier
Yuqian Zhang \and
Shuai Liu \and
% BLR
Jie Liu \and
Zhuhao Zhang \and
% Nankai-CVLab
Lishen Qu \and
Zhihao Liu \and
Shihao Zhou \and
Yaqi Luo \and
Juncheng Zhou \and
Jufeng Yang \and
% MagicBlur
Qianfeng Yang \and
Qiyuan Guan \and
Xiang Chen \and
Guiyue Jin \and
Jiyu Jin
}
\maketitle
\let\thefootnote\relax\footnotetext{
$^\dagger$
George Ciubotariu, Florin-Alexandru Vasluianu, Zhuyun Zhou, Nancy Mehta, and Radu Timofte are the challenge organizers. The other authors participated in the challenge.
Each team described its own method in the report, shortened by the organizers to meet 8 page criteria. Appendix~\ref{sec:teams} contains the teams, affiliations, and architectures if available. \\ AIM 2025 webpage:~\url{https://cvlai.net/aim/2025}.\\
Project Page:~\url{https://github.com/george200150/MIORe}.
} 
\begin{abstract}
This paper presents a comprehensive review of the AIM 2025 High FPS Non-Uniform Motion Deblurring Challenge, highlighting the proposed solutions and final results. The objective of this challenge is to identify effective networks capable of producing clearer and visually compelling images in diverse and challenging conditions, by learning representative visual cues for complex aggregations of motion types. A total of 68 participants registered for the competition, and 9 teams ultimately submitted valid entries. This paper thoroughly evaluates the state-of-the-art advances in high-FPS single image motion deblurring, showcasing the significant progress in the field, while leveraging samples of the novel dataset, MIORe, that introduces challenging examples of movement patterns.
\end{abstract}

\section{Introduction}
\label{sec:introduction}

Blur refers to the loss of sharpness in an image, characterized by the smoothing of fine details and edges due to the attenuation of high spatial frequencies. The two main types of blur approached by the computer vision community are \emph{Defocus Blur} and \emph{Motion Blur}.

\emph{Motion Blur} is the streaking or smearing of moving objects (or the background when the camera moves) captured during the finite exposure time of an imaging sensor. The reason is that, in digital photography, the camera sensor integrates light over a period during which scene elements change position, effectively convolving the latent sharp image with a motion‐dependent point spread function.

\emph{Single Image Motion Deblurring} (SIMD) is dealing with scenes in which, due to the rapid motion of either the ego camera or the scene entities, the sharpness and clarity are lost. The task aims to mitigate the degradations from one single input image which, although ill-posed, represents a highly popular task within the research community.

Given an RGB image degraded by motion blur (LQ) and a high-quality sharp ground-truth RGB image (HQ), the task aims to recover the sharpness of HQ from LQ by leveraging motion pattern recognition and removal.

%% cross-referencing AIM 2025 associated challenges
This challenge is one of the AIM 2025~\footnote{\url{https://www.cvlai.net/aim/2025/}} workshop associated challenges on: high FPS non-uniform motion deblurring~\cite{aim2025highfps}, rip current segmentation~\cite{aim2025ripseg}, inverse tone mapping~\cite{aim2025tone}, robust offline video super-resolution~\cite{aim2025videoSR}, low-light raw video denoising~\cite{aim2025videodenoising}, screen-content video quality assessment~\cite{aim2025scvqa}, real-world raw denoising~\cite{aim2025rawdenoising}, perceptual image super-resolution~\cite{aim2025perceptual}, efficient real-world deblurring~\cite{aim2025efficientdeblurring}, 4K super-resolution on mobile NPUs~\cite{aim20254ksr}, efficient denoising on smartphone GPUs~\cite{aim2025efficientdenoising}, efficient learned ISP on mobile GPUs~\cite{aim2025efficientISP}, and stable diffusion for on-device inference~\cite{aim2025sd}. Descriptions of the datasets, methods, and results can be found in the corresponding challenge reports.

\section{Challenge Data}

The challenge utilizes two newly introduced datasets, MIORe and VAR-MIORe~\cite{ciubotariu2025miore}, both designed to capture complex motion patterns and diverse blur scenarios. As specified on the Codabench competition page~\cite{Codabench}, the proposed data splits contain challenging blurry images synthetically generated from sharp ground-truth frames recorded with a high-frame-rate CHRONOS 2.1-HD camera~\cite{chronos}.

MIORe emphasizes diverse and realistic motion patterns, incorporating mild to moderate motion blur along with occasional defocus blur. In contrast, VAR-MIORe extends this design toward extreme motion conditions, simulating scenarios with blur up to four times stronger than in MIORe. Together, the datasets provide a comprehensive testbed spanning both moderate and extreme blur regimes.  

\section{Tracks}

\subsection{Track 1: Moderate Motion Blur}
Track~1 leverages MIORe to evaluate deblurring methods under blur intensities comparable to existing benchmarks. The key novelty lies in the diversity of motion patterns: although blur magnitude varies within each scene, values are normalized via precomputed mean and maximum optical flow statistics. This ensures consistent difficulty while retaining scene variability. The dataset also introduces defocus blur to approximate real-world capture conditions.  

The training split contains 7{,}860 paired samples (input and ground truth) at full-HD resolution (1920$\times$1080, 8-bit RGB). The validation split consists of 80 images, and the testing split includes 75 images. Ground-truth labels are available only for the training set.  

\subsection{Track 2: High-Motion Blur}
Track~2 is built from VAR-MIORe and focuses on extreme motion scenarios largely unexplored in prior work. The captured sequences exhibit blur intensities ranging from above average to up to four times the severity of Track~1. This setting is designed to benchmark robustness under severe motion degradation, posing a more challenging restoration problem.  

The training split consists of 1{,}299 paired samples, while the validation and testing splits contain 67 and 63 images, respectively, following the same resolution and format as Track~1.

\section{Evaluation}

This challenge was a double-track competition with the same ranking criteria for both tracks. The following are the evaluation criteria:
\begin{itemize}
    \item The reconstruction fidelity in terms of PSNR;
    \item The Structured Similarity Index (SSIM) \cite{wang2004image} score;
    \item The LPIPS \cite{zhang2018perceptual} distance between restorations and the ground-truth images. We used ImageNet pre-trained AlexNet \cite{10.1145/3065386} for LPIPS feature extraction.
\end{itemize}

\section{Challenge phases}

\begin{enumerate}
    \item \textbf{Development phase:} In this phase, participants were given access to the description of the challenge, together with a set of 7860 pairs of images for Track 1 and 1299 for Track 2 to train their models.
    \item \textbf{Validation phase:} During this phase, to validate the solutions, participants received a set of 80 input images for Track 1 and 67 for Track 2 from the validation split, with the ground truths not shared. To enable measurement of validation performance, a Codabench server \cite{Codabench} was set up, comparing submitted images uploaded by each participant with private reference images. 
    \item \textbf{Final phase:} Sets of 75 input images for Track 1 and 63 for Track 2 from the testing split were sent to challenge participants, with evaluation performed by the Codabench server. The fine-tuning of the test set was restricted by limiting user submissions. Finally, participants received a submission template, with instructions for the final preparation of the submission.
    Each team provided their method description, corresponding code, information regarding team members, affiliations, and the final set of restored images, corresponding to the input from the testing split.
    
\end{enumerate}

\section{Challenge Results}

The challenge ended with a total of 9 valid submissions.
Section \ref{sec:methods_and_teams} provides details of each solution ranked in the final phase.

\begin{figure*}[t]
    \centering
    \includegraphics[width=.99\linewidth]{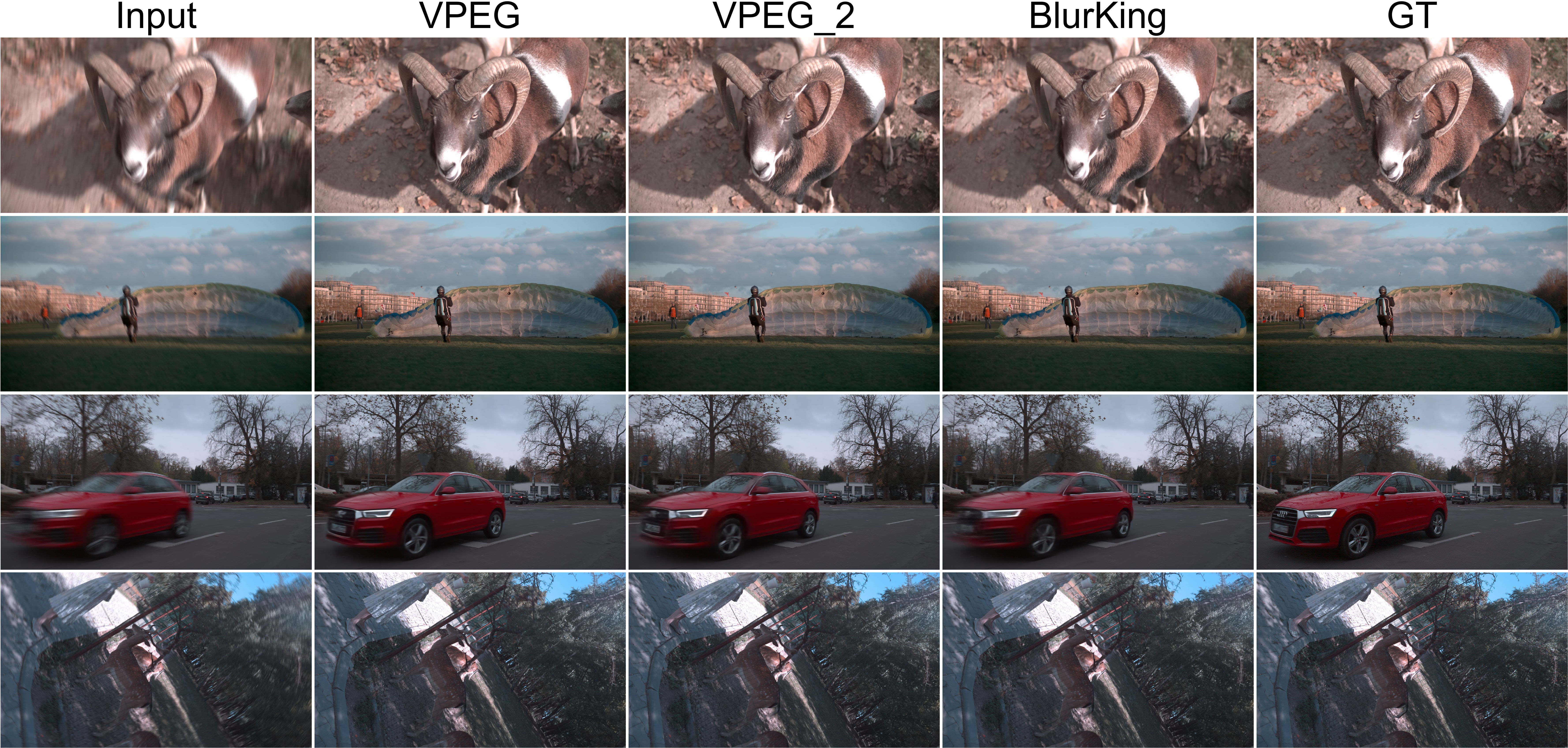}
    \caption{
    Visual comparison of the solutions proposed by top teams on samples from the AIM 2025 High FPS Motion Deblurring Challenge testing split of
    Track 1.
    }
    \label{fig:aim25_results_1}
\end{figure*}

\begin{figure*}[t]
    \centering
    \includegraphics[width=.99\linewidth]{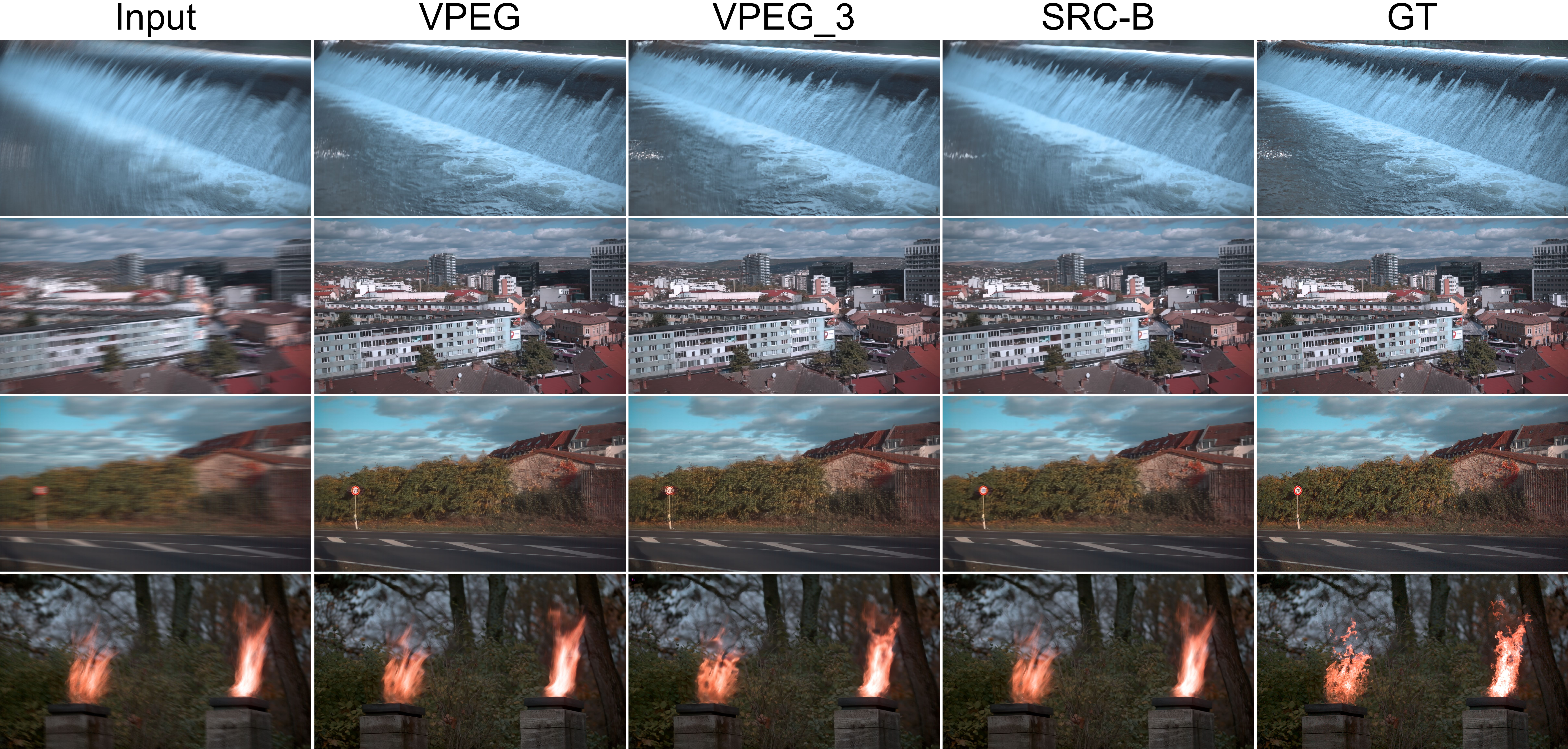}
    \caption{
    Visual comparison of the solutions proposed by top teams on samples from the AIM 2025 High FPS Motion Deblurring Challenge testing split of
    Track 2.
    }
    \label{fig:aim25_results_2}
\end{figure*}

The quantitative results for the two respective tracks are reported in \cref{tbl:aim25_results_1} and \cref{tbl:aim25_results_2}. The submitted approaches demonstrate strong performance in terms of perceptual quality metrics. In addition, qualitative comparisons, as shown in \cref{fig:aim25_results_1} and \cref{fig:aim25_results_2}, further substantiate the effectiveness of solutions through visual inspection.

\begin{table*}[t]
\footnotesize
\centering
\caption{Evaluation and Rankings in the AIM 2025 High FPS Motion Deblurring Challenge - Track 1. The ``rk."s indicate the respective standings of teams based on their performances in different metrics on the challenge's test dataset. The ``Final Rank" represents a composite metric, derived from the average ranking.}
\begin{tabular}{l|ccc|ccc|ccc|c}
\toprule
Team & PSNR $\uparrow$ & SSIM $\uparrow$ & LPIPS $\downarrow$ & rk. PSNR & rk. SSIM & rk. LPIPS & Params. (M) & Runtime (s) & Device & Final Rank \\
\midrule
VPEG          & 34.484        & 0.9026        & 0.1386         & 1                  & 1                  & 2 & 17.1 & 52 & 3090                   & 1                   \\
VPEG\_2       & 34.155        & 0.8990        & 0.1431         & 2                  & 2                  & 3 & 14.9 & 2.1 & 4090                   & 2                   \\
BlurKing      & 33.337        & 0.8870        & 0.1640         & 3                  & 3                  & 4 & 15 & 1 & 4090                   & 3                   \\
SRC-B       & 33.185        & 0.8833        & 0.1745         & 4                  & 4                  & 5 & 35 & 0.5 & A100                   & 4                   \\
X-L           & 32.627        & 0.8757        & 0.1844         & 5                  & 5                  & 6 & 5 & 0.5 & A100                   & 5                   \\
Mier          & 28.807        & 0.7800        & 0.1295         & 9                  & 9                  & 1 & 12k & 100 & A100                   & 6                   \\
BLR           & 32.387        & 0.8727        & 0.1881         & 6                  & 6                  & 7 & 49 & 5 & A100                   & 6                   \\
Nankai-CVLab  & 32.088        & 0.8670        & 0.1905         & 7                  & 7                  & 8 & 24 & 3.6 & V100                   & 8                   \\
MagicBlur     & 31.535        & 0.8565        & 0.2129         & 8                  & 8                  & 9 & 26.1 & 15.9 & vGPU                   & 9                  \\

\bottomrule
\end{tabular}
\label{tbl:aim25_results_1}
\end{table*}

\begin{table*}[t]
\footnotesize
\centering
\caption{Evaluation and Rankings in the AIM 2025 High FPS Motion Deblurring Challenge - Track 2. The ``rk."s indicate the respective standings of teams based on their performances in different metrics on the challenge's test dataset. The ``Final Rank" represents a composite metric, derived from the average ranking.}

\begin{tabular}{l|ccc|ccc|ccc|c}
\toprule
Team & PSNR $\uparrow$ & SSIM $\uparrow$ & LPIPS $\downarrow$ & rk. PSNR & rk. SSIM & rk. LPIPS & Params. (M) & Runtime (s) & Device & Final Rank \\
\midrule
VPEG               & 30.287        & 0.8434        & 0.2388         & 1                  & 1                  & 2 & 17.1 & 52 & 3090                   & 1                   \\
VPEG\_3            & 28.582        & 0.8096        & 0.2661         & 2                  & 2                  & 3 & 14.9 & 2.1 & 4090                   & 2                   \\
SRC-B            & 28.479        & 0.8046        & 0.3035         & 3                  & 3                  & 4 & 35 & 0.5 & A100                   & 3                   \\
X-L                & 27.474        & 0.7885        & 0.3284         & 4                  & 4                  & 5 & 5 & 0.5 & A100                   & 4                   \\
Mier               & 25.898        & 0.7182        & 0.2005         & 8                  & 8                  & 1 & 12k & 100 & A100                   & 5                   \\
Nankai-CVLab       & 27.190        & 0.7851        & 0.3412         & 6                  & 5                  & 6 & 24 & 3.6 & V100                   & 5                   \\
Sharper            & 27.260        & 0.7768        & 0.3575         & 5                  & 6                  & 7 & 15 & 1 & 4090                   & 7                   \\
MagicBlur          & 26.570        & 0.7647        & 0.3808         & 7                  & 7                  & 8 & 26.1 & 15.9 & vGPU                   & 8                  \\
\bottomrule
\end{tabular}
\label{tbl:aim25_results_2}

\end{table*}

\section{Challenge Methods and Teams}
\label{sec:methods_and_teams}

\subsection{VPEG}
\begin{figure*}[htp]
    \centering
    \includegraphics[width=0.95\linewidth]{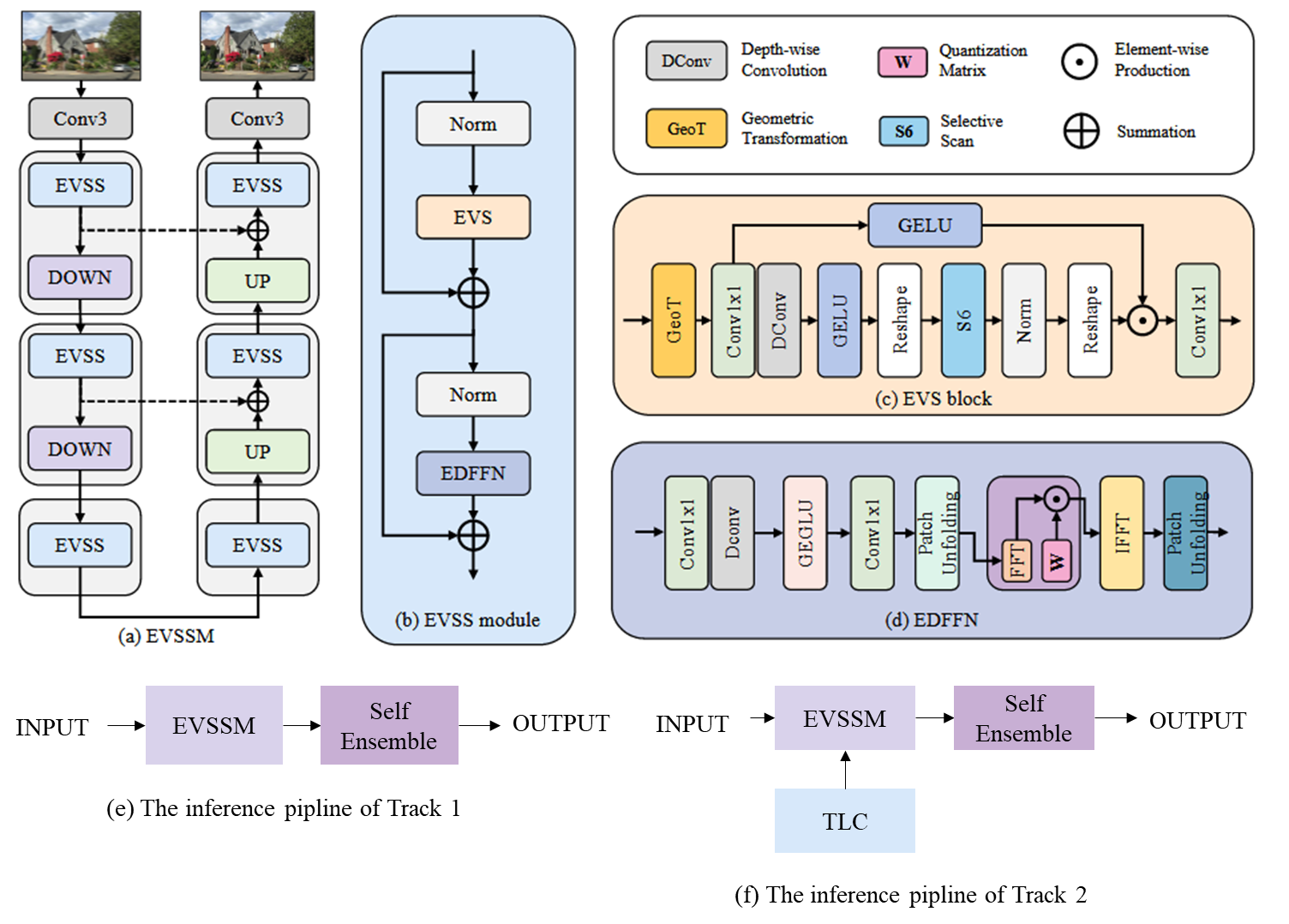}
    \caption{EVSSM architecture and inferences piplines}
    \label{fig:EVSSM}
\end{figure*}

The team employed the Efficient Visual State Space Models (EVSSM) architecture~\cite{EVSSM}: the EVS block alternates transposition or flipping operations before feature processing to preserve the spatial structure of images, while the (efficient discriminative frequency domain-based feedforward network) EDFFN block filters high-frequency information in the frequency domain to improve detail recovery and reduce computational costs.
\\
Figure \ref{fig:EVSSM} shows the detail architecture of the used EVSSM~\cite{EVSSM}. 
The team participated in both Track 1 and Track 2 of the AIM competition with the same model structure. The differences between the two approaches lie in the multi-stage training processes and inference piplines, which are elaborated in the following.
\\
\noindent \textbf{Multi-stage training strategy.}
For Track 1, VPEG used the same model parameter settings as in the original EVSSM~\cite{EVSSM} and a multi-stage training strategy to improve the deblur performance on the given high FPS motion blur dataset. 
Stage I: the the employed EVSSM was trained on the GoPro dataset~\cite{nah2017deep} with input patch of 128$\times$128 pixels and the mini-batch is 64.
Stage II: then, the team trained the model on the competition dataset~\cite{ciubotariu2025miore} with an input patch of 128$\times$128 pixels and the mini-batch 64.
Stage III: next, the selected competition dataset~\cite{ciubotariu2025miore} was used for training, with an input patch of 256$\times$256 pixels and 8 samples per mini-batch.
Stage IV: lastly, the team trained the employed EVSSM on the selected competition dataset~\cite{ciubotariu2025miore} with input patch of 288$\times$288 pixels and the mini-batch of 4. 
\\
For Track 2, VPEG also adopted the training strategy of gradually increasing patch size, while leveraging the pre-trained model from Track 1 Stage 3.
\\
In terms of data augmentation, the team randomly applied methods such as gamma correction and brightness adjustment to enhance the model's ability to recover details in exposure and low-light scenarios.
\\

\subsection{VPEG\_2\_3}
\begin{figure*}[htp]
    \centering
    \includegraphics[width=0.95\linewidth]{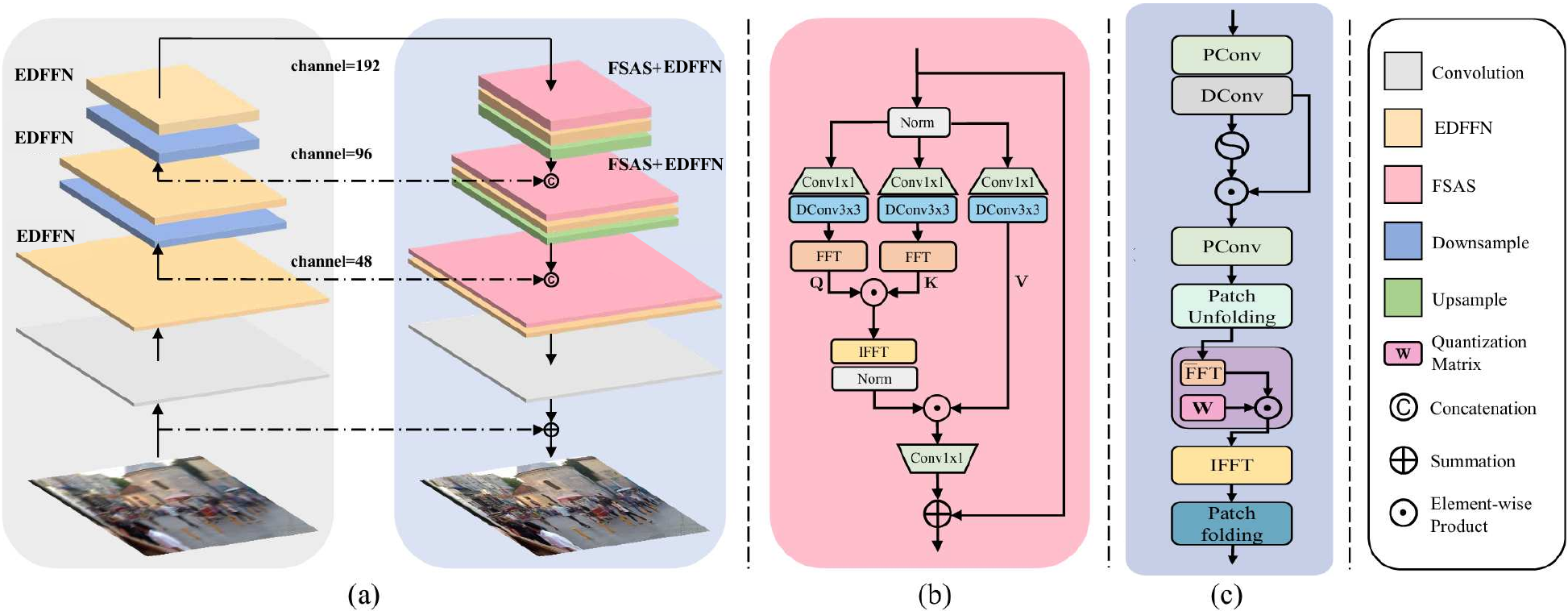}
    \caption{FFTformer+ architectures. (a) The proposed asymmetric encoder-decoder network that only contains EDFFN in the encoder module and both FSAS and EDFFN in the decoder module for image deblurring. (b) The FSAS module. (c) The EDFFN module.}
    \label{fig:FFTformer+}
\end{figure*}

VPEG\_2 and VPEG\_3 propose an efficient frequency domain-based transformer and discriminative FFN for High-Quality image deblurring method named FFTformer+.
FFTformer+ adopt the frequency-domain attention mechanism proposed in FFTformer~\cite{Fftformer}, namely the Frequency domain-based Self-Attention Solver (FSAS), to perform efficient attention computation. 
FSAS significantly reduces the computational cost while preserving the effectiveness of attention, enabling the accurate reconstruction of image structures. 
In addition, FFTformer+ integrates the EDFFN of EVSSM~\cite{EVSSM} to selectively filter frequency-domain features, facilitating efficient aggregation of local information.
Figure \ref{fig:FFTformer+} provides an overview of the proposed FFTformer+.

\noindent \textbf{Training strategy.}
The AdamW optimizer is employed with default parameters during training.
Data augmentation methods include flipping and rotation operations to generate training data.
The framework employs progressive training similar to but simpler than~\cite{Restormer}: the training starts with the patch size of 128 $\times$ 128 pixels and the batch size of 64 for 300,000 iterations, where the learning rate gradually reduces from $1 \times 10^{-3}$ to $1 \times 10^{-7}$. 
Then the patch size is enlarged to $256 \times 256$ pixels with 16 samples per batch for 300,000 iterations where the learning rate is initialized as $5 \times 10^{-4}$ and decreases to $1 \times 10^{-7}$.
The learning rate is updated based on the Cosine Annealing scheme~\cite{cosine}.
\subsection{SRC-B}
 
\begin{figure*}[htbp]
	\centering
	\includegraphics[width=0.95\linewidth]{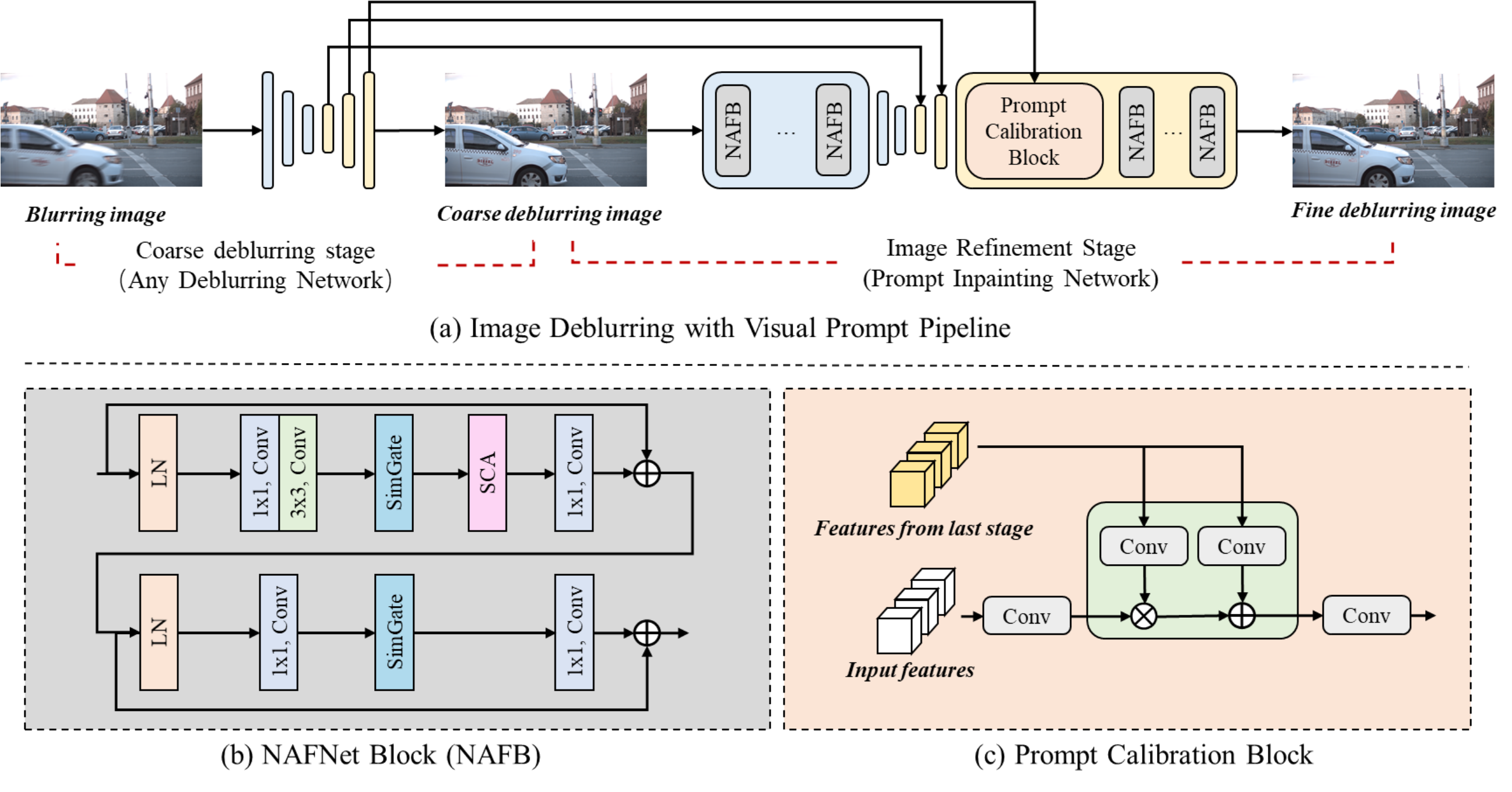}
	\caption{Overview of the technique proposed by team SRC-B for deblurring.}
	\label{SRC-B_net}
\end{figure*}

The team proposes a visual prompt-based image deblurring framework.
The method adopts a two-stage pipeline: (1) coarse deblurring and (2) refined adjustment. Both stages employ identical architectures based on NAFNet~\cite{NAFNet}, a lightweight restoration network designed without nonlinear activation functions on its main computational paths. This omission facilitates the preservation of high-frequency information, which is critical for recovering fine image details in deblurring tasks. The NAFNet block serves as the fundamental feature extraction unit in both stages.  
Figure \ref{SRC-B_net} provides an overview of the proposed architecture.

In the first stage, the blurred image is processed to remove the majority of motion blur artifacts, producing a preliminary sharp output. Features extracted during this stage are treated as visual prompts, encoding contextual and structural information. The second stage consumes both the preliminary result and these visual prompts to refine residual blur, enhance texture fidelity, and restore structural integrity. This progressive design allows the network to take advantage of semantic cues from the first stage to achieve high-fidelity reconstructions.  

\textbf{Training details.} The model is optimized with Adam, using an initial learning rate of $2\times10^{-4}$ over 600,000 iterations. The objective function is the L1 loss, which penalizes the pixel-wise absolute differences between the restored and ground-truth images, effectively preserving sharp details without excessive smoothing. This training configuration, coupled with the two-stage prompt-guided refinement, enables high FPS motion deblurring while maintaining competitive restoration quality across diverse blur conditions.

\subsection{X-L}

\begin{figure*}[htbp]
	\centering
	\includegraphics[width=0.95\linewidth]{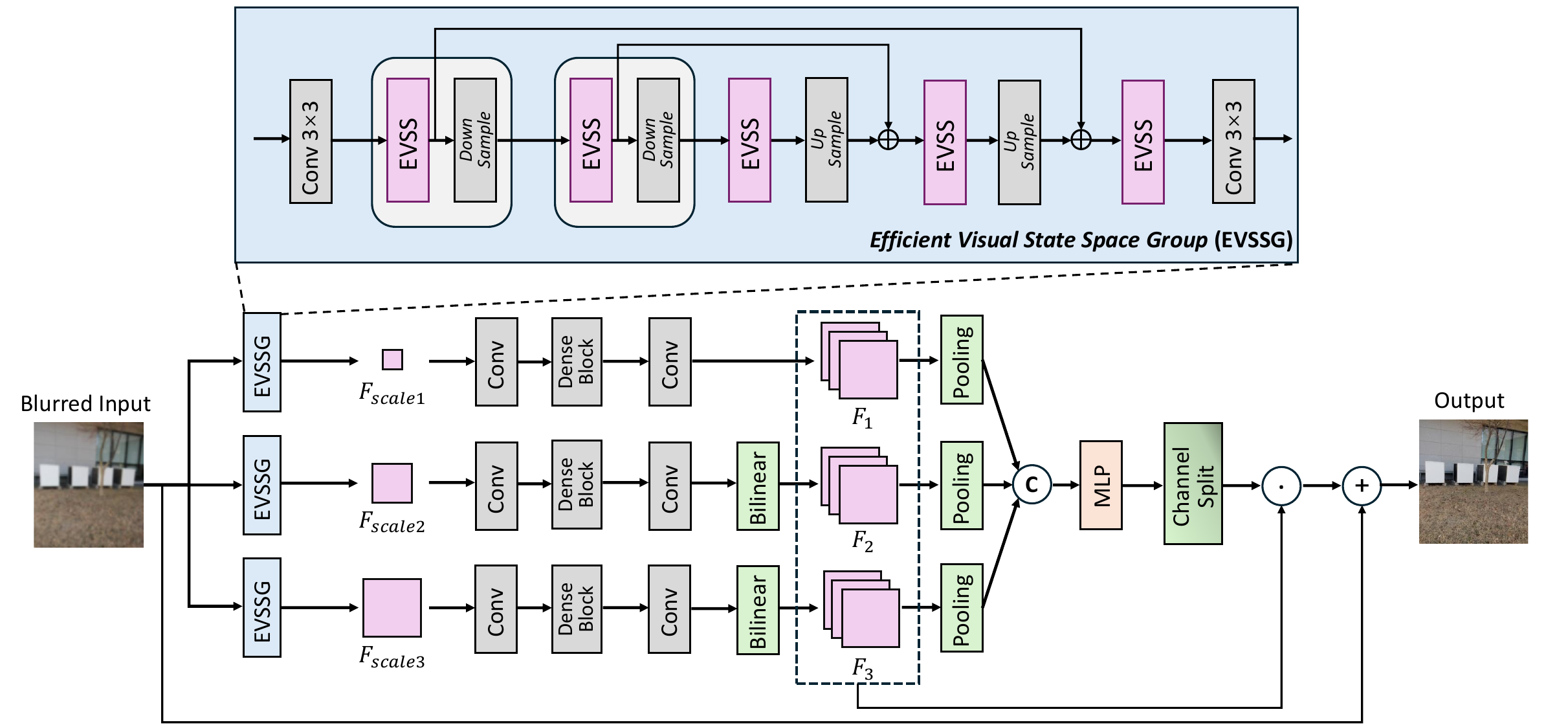}
	\caption{Overview of the technique proposed by team X-L for deblurring.}
	\label{X-L_net}
\end{figure*}

The team introduces a deblurring framework that integrates EVSSM~\cite{EVSSM} with a semantically-aligned, scale-aware fusion strategy~\cite{ESDNet}, as illustrated in Figure~\ref{X-L_net}. The EVSSM backbone provides efficient long-range dependency modeling, enabling effective handling of non-uniform motion blur in high-frame-rate (HFR) sequences.  

To address multi-scale feature interaction, the architecture incorporates a dynamic weighting mechanism, where an MLP predictor estimates blending weights for different scale branches. This ensures semantically-aligned integration of multi-scale features, allowing the model to adaptively emphasize spatial details most relevant to the underlying scene content. Blurred frames are first processed through EVSSM-based multi-scale encoders. The resulting features are fused via the scale-aware module and subsequently refined through cascaded EVSSM blocks before reconstruction. This design effectively captures spatially varying blur patterns while maintaining temporal coherence, achieving robust deblurring under real-time constraints.  

\textbf{Training details.} The training follows the EVSSM~\cite{EVSSM} protocol, initialized from a pre-trained model. The network is trained for 100 epochs (12 hours) on an NVIDIA A100 GPU. This configuration balances accuracy and efficiency, enabling deployment in HFR video deblurring scenarios.

\subsection{BlurKing\_Sharper}
\begin{figure}[htbp]
	\centering
	\includegraphics[width=\linewidth]{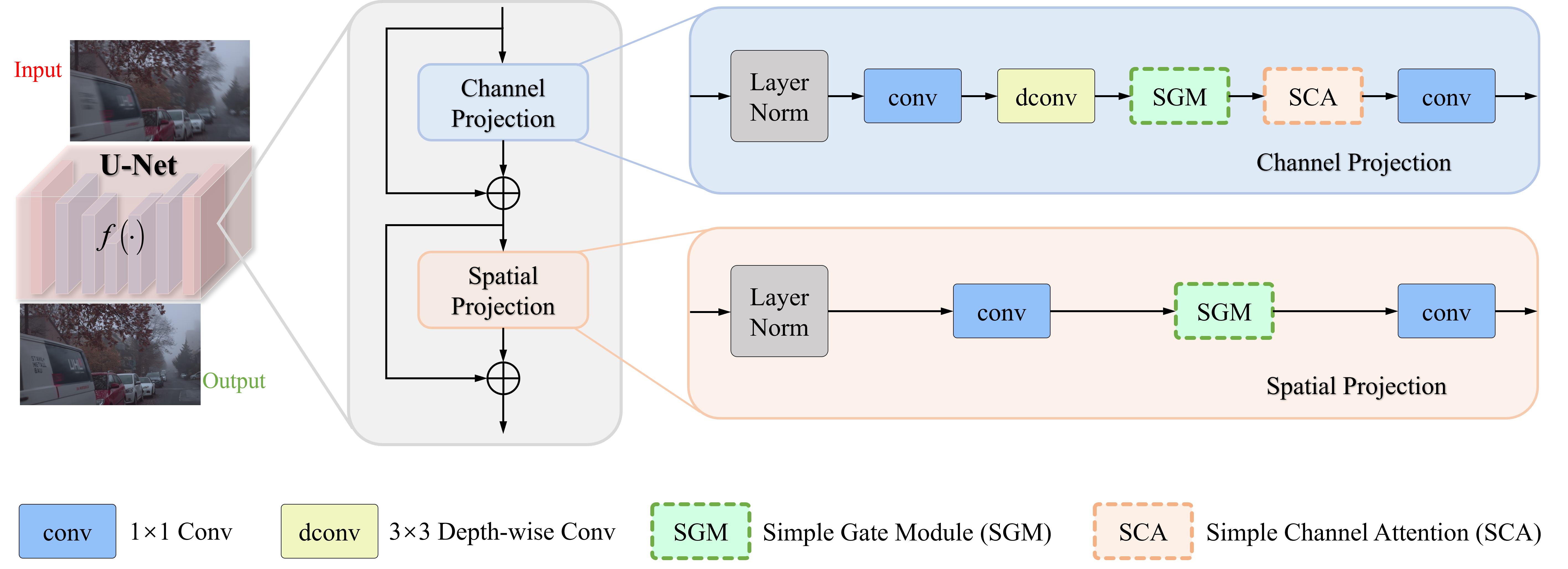}
	\caption{Overview of the technique proposed by team BlurKing / Sharper for deblurring.}
	\label{BlurKing_Sharper_net}
\end{figure}

The Motion Deblur network \cite{Lu2025Deblurring} is primarily composed of modules such as Gated Linear Units, SimpleGate, and Simplified Channel Attention, inspired by references \cite{chen2022simple,Lu_2025_CVPR_AALN,Lu_2025_CVPR_EvenFormer, Lu_2024_CVPR_HirFormer}. The integration of these intricate components results in a robust baseline capable of effectively addressing high-resolution Motion Deblur tasks.
In order to ensure the stability of the training under a large amount of data, inspired by reference \cite{Lu_2025_CVPR_ILAWR,Lu2025Tone,lu2025elucidating}, the team adopts a progressive multiscale training method to further enhance the performance of the network.
The architecture is shown in Figure \ref{BlurKing_Sharper_net}.

\subsection{Mier}

The team developed \emph{EasyControl-Lite}, a generative restoration framework derived from EasyControl~\cite{zhang2025easycontrol} and FLUX~\cite{labs2025flux1kontextflowmatching}, trained on the provided datasets. Mixed training was performed using data from both Track~1 and Track~2, with equal weighting applied to their respective test sets. To address out-of-memory issues arising from high-resolution training, the conditional branch concatenation was replaced with an additive integration into the noise branch.  

As a generative model, the method prioritizes perceptual quality over pixel-wise fidelity, leading to moderate PSNR scores but competitive LPIPS performance and strong subjective visual results. The architecture balances generative flexibility with efficient conditioning, enabling high-quality reconstructions within the challenge constraints.  

\textbf{Training details.} The model was initialized from the FLUX pre-trained weights (12B parameters) and trained for 20~epochs (25~hours) on an NVIDIA A100 GPU. This setup preserved the strengths of the base generative backbone while adapting it to the multi-track training scenario.

\subsection{BLR}

The team adopts a Transformer-style architecture~\cite{LoFormer}, leveraging pre-trained model weights from three large-scale datasets, which are GoPro~\cite{GoPro}, RealBlurJ, and RealBlurR~\cite{RealBlur}. These weights are fused through a weighted averaging strategy to integrate complementary priors from diverse blur scenarios. The fused model is then fine-tuned on the official competition training dataset to adapt to the specific motion blur characteristics of the challenge.  

An additional experiment combines an attention-based model with a diffusion model to handle distinct dynamic subjects within the same image; however, this hybrid approach does not yield performance improvements over the primary method.  

\textbf{Training details.} The fusion step is followed by a warm-up phase of 5,000 iterations and a main training phase of 100,000 iterations. Optimization is performed with a learning rate of $1\times10^{-4}$ and a batch size of 8 on 4~NVIDIA A100 GPUs. This procedure effectively consolidates multi-dataset knowledge while tailoring the model for the deblurring tasks of the competition.

\subsection{Nankai-CVLab}
\begin{figure}
    \centering
    \includegraphics[width=\linewidth]{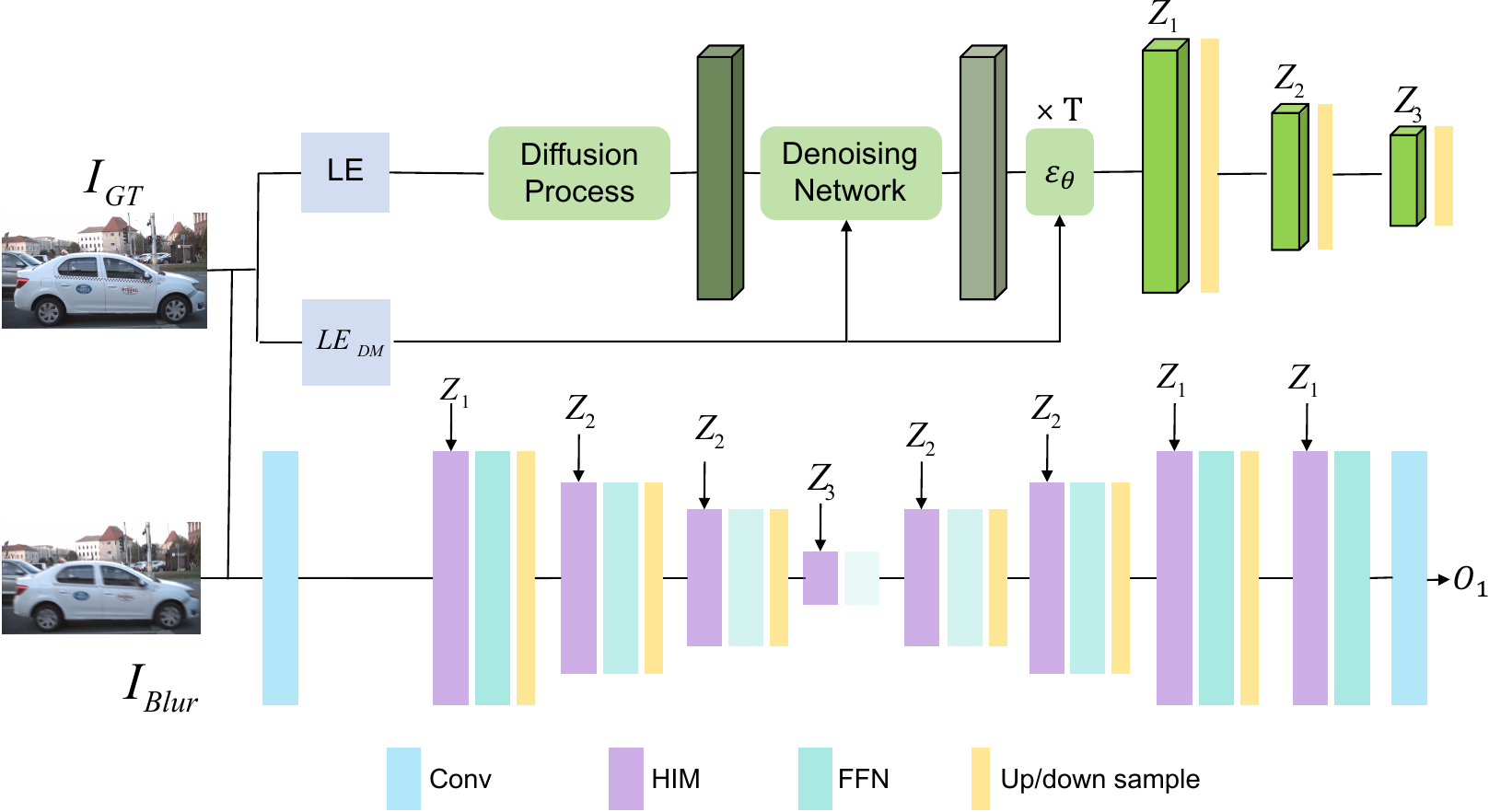}
    \caption{Details of the Nankai-CVLab team's architectural approach to deblurring.}
    \label{fig:model}
\end{figure}

Nankai-CVLab introduced RecuDiff-UT, a two-stage diffusion-Transformer framework based on~\cite{chen2024hierarchical} and~\cite{ciubotariu2025miore} for artifact-free recursive inference via a self-corrective training paradigm. By integrating a task-specific encoder with a U-Transformer diffusion backbone as shown in Fig. \ref{fig:model}, the team aims to address performance degradation in conventional transformer models~\cite{liu2021swin} under iterative resampling, explicitly preventing error accumulation while optimizing reconstruction fidelity.

\noindent \textbf{Hybrid Architecture Training.}
The framework employs a multi-pass inference pipeline beginning with a pre-trained task-specific encoder that maps input features to a latent space representation. With the encoder parameters fixed, the U-Transformer backbone then generates initial output $O_1$ through diffusion-guided feature denoising, where progressively refined noise estimates from the diffusion process stage-wise condition the transformer's restoration path. This output $O_1$ is recursively fed back into the same model for self-corrective refinement. The joint optimization enforces: (a) reconstruction fidelity against ground truth via $L_1$ and perceptual losses, (b) inter-pass consistency across recursive outputs, and (c) latent space stability through explicit architectural regularization, ultimately converging to artifact-free results.
\\
\noindent \textbf{Recursive Stability Loss.}
The total loss integrates standard reconstruction ($L_1$) and perceptual ($L_{\text{vgg}}$) terms with a novel stability objective $L_{\text{stable}}$. This core component enforces consistency between consecutive recursive outputs: when initial prediction $O_1 = f_\theta(x)$ is fed back into the model, the team minimizes the divergence of the second output $O_2 = f_\theta(O_1)$ from $O_1$ using a composite metric:
\begin{equation}
L_{\text{stable}} = \alpha \cdot \text{LPIPS}(O_1, O_2) + \beta \cdot (1 - \text{SSIM}(O_1, O_2))
\label{eq:stable_loss}
\end{equation}
where $\alpha,\beta$ balance structural and perceptual similarity. The complete objective $L_{\text{total}} = L_1 + \lambda_1 L_{\text{vgg}} + \lambda_2 L_{\text{stable}}$ explicitly penalizes output drift across recursion cycles while maintaining reconstruction fidelity, ensuring stable multi-pass inference without artifact accumulation.  
\subsection{MagicBlur}

\begin{figure}[htbp]
	\centering
	\includegraphics[width=\linewidth]{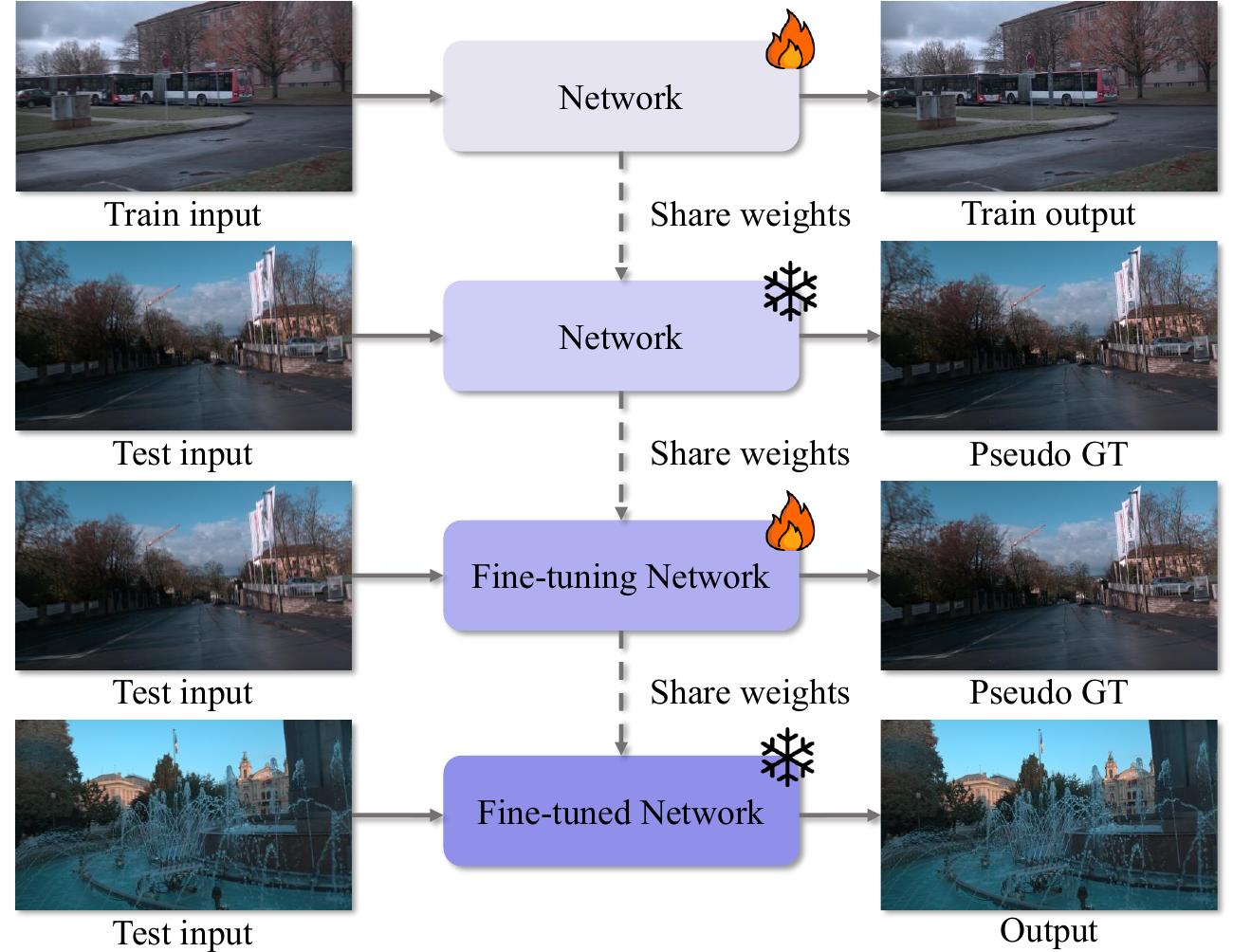}
	\caption{Overview of the technique proposed by team MagicBlur for deblurring.}
	\label{MagicBlur_net}
\end{figure}

MagicBlur adopts a fine-tuning strategy using pseudo ground truth images to address this challenge. The overall technical pipeline is illustrated in Fig. \ref{MagicBlur_net}. Since the ground truth for the test images is not publicly available, this team first pretrains the model using Restormer~\cite{zamir2022restormer} on the training set \cite{ciubotariu2025miore} in the first stage. Next, this team processes the test set inputs with the pretrained shared weights to generate pseudo ground truth images for the test set. In the second stage, this team fine-tunes the model on the test set using the pseudo ground truth pairs, based on the pretrained weights from the first stage. This approach provides a clear direction for transferring pretrained knowledge. Subsequently, this team applies preprocessing operations such as rotation and scaling transformations to the input images, then processes them using the fine-tuned shared weights. Finally, this team applies a median filtering technique to produce the final output.

\section{Conclusion}

We introduced the first edition of our AIM 2025 High FPS Motion Deblurring challenge, which enjoyed significant attention from the computer vision community. The challenge was split into two tracks: the first one dealing with reasonable blurry samples, that have been previously analyzed in the single image motion deblurring literature, and a second track designed for extreme conditions, pushing the boundaries of what models could perform. Both tracks received about the same number of submissions from teams that proposed their single solution for both scenarios. We presented the proposed models of the participants and quantified the performance in terms of PSNR, SSIM and LPIPS, characterizing the restored images, which were the criteria for ranking the both tracks of the challenge.

\section*{Acknowledgments}
This work was partially supported by the Alexander von Humboldt Foundation. We thank the AIM 2025 sponsors: AI Witchlabs and University of W\"urzburg (Computer Vision Lab).

{\small
\bibliographystyle{ieee_fullname}
\bibliography{egbib}

\begin{thebibliography}{10}\itemsep=-1pt

\bibitem{NAFNet}
Liangyu Chen, Xiaojie Chu, Xiangyu Zhang, and Jian Sun.
\newblock Simple baselines for image restoration.
\newblock In Shai Avidan, Gabriel~J. Brostow, Moustapha Ciss{\'{e}}, Giovanni~Maria Farinella, and Tal Hassner, editors, {\em Computer Vision - {ECCV} 2022 - 17th European Conference, Tel Aviv, Israel, October 23-27, 2022, Proceedings, Part {VII}}, volume 13667 of {\em Lecture Notes in Computer Science}, pages 17--33. Springer, 2022.

\bibitem{chen2022simple}
Liangyu Chen, Xiaojie Chu, Xiangyu Zhang, and Jian Sun.
\newblock Simple baselines for image restoration.
\newblock In {\em European conference on computer vision}, pages 17--33. Springer, 2022.

\bibitem{chen2024hierarchical}
Zheng Chen, Yulun Zhang, Ding Liu, Jinjin Gu, Linghe Kong, Xin Yuan, et~al.
\newblock Hierarchical integration diffusion model for realistic image deblurring.
\newblock {\em NeurIPS}, 2024.

\bibitem{aim2025highfps}
George Ciubotariu, Florin-Alexandru Vasluianu, Zhuyun Zhou, Nancy Mehta, Radu Timofte, et~al.
\newblock {AIM} 2025 high {FPS} non-uniform motion deblurring challenge report.
\newblock In {\em Proceedings of the IEEE/CVF International Conference on Computer Vision (ICCV) Workshops}, 2025.

\bibitem{ciubotariu2025miore}
George Ciubotariu, Zhuyun Zhou, Zongwei Wu, and Radu Timofte.
\newblock {MIOR}e \& {VAR-MIOR}e: Benchmarks to push the boundaries of restoration.
\newblock In {\em Proceedings of the IEEE International Conference on Computer Vision (ICCV)}. {IEEE} Computer Society, 2025.

\bibitem{aim2025ripseg}
Andrei Dumitriu, Florin Miron, Florin Tatui, Radu~Tudor Ionescu, Radu Timofte, Aakash Ralhan, Florin-Alexandru Vasluianu, et~al.
\newblock {AIM} 2025 rip current segmentation ({RipSeg}) challenge report.
\newblock In {\em Proceedings of the IEEE/CVF International Conference on Computer Vision (ICCV) Workshops}, 2025.

\bibitem{aim2025efficientdeblurring}
Daniel Feijoo, Paula Garrido, Marcos Conde, Jaesung Rim, Alvaro Garcia, Sunghyun Cho, Radu Timofte, et~al.
\newblock Efficient real-world deblurring using single images: {AIM} 2025 challenge report.
\newblock In {\em Proceedings of the IEEE/CVF International Conference on Computer Vision (ICCV) Workshops}, 2025.

\bibitem{aim20254ksr}
Andrey Ignatov, Georgy Perevozchikov, Radu Timofte, et~al.
\newblock {4K} image super-resolution on mobile {NPUs}: {Mobile AI \& AIM 2025} challenge report.
\newblock In {\em Proceedings of the IEEE/CVF International Conference on Computer Vision (ICCV) Workshops}, 2025.

\bibitem{aim2025sd}
Andrey Ignatov, Georgy Perevozchikov, Radu Timofte, et~al.
\newblock Adapting stable diffusion for on-device inference: {Mobile AI \& AIM 2025} challenge report.
\newblock In {\em Proceedings of the IEEE/CVF International Conference on Computer Vision (ICCV) Workshops}, 2025.

\bibitem{aim2025efficientdenoising}
Andrey Ignatov, Georgy Perevozchikov, Radu Timofte, et~al.
\newblock Efficient image denoising on smartphone {GPUs}: {Mobile AI \& AIM 2025} challenge report.
\newblock In {\em Proceedings of the IEEE/CVF International Conference on Computer Vision (ICCV) Workshops}, 2025.

\bibitem{aim2025efficientISP}
Andrey Ignatov, Georgy Perevozchikov, Radu Timofte, et~al.
\newblock Efficient learned smartphone {ISP} on mobile {GPUs}: {Mobile AI \& AIM 2025} challenge report.
\newblock In {\em Proceedings of the IEEE/CVF International Conference on Computer Vision (ICCV) Workshops}, 2025.

\bibitem{aim2025videoSR}
Nikolai Karetin, Ivan Molodetskikh, Dmitry Vatolin, Radu Timofte, et~al.
\newblock {AIM} 2025 challenge on robust offline video super-resolution: Dataset, methods and results.
\newblock In {\em Proceedings of the IEEE/CVF International Conference on Computer Vision (ICCV) Workshops}, 2025.

\bibitem{Fftformer}
Lingshun Kong, Jiangxin Dong, Jianjun Ge, Mingqiang Li, and Jinshan Pan.
\newblock Efficient frequency domain-based transformers for high-quality image deblurring.
\newblock In {\em CVPR}, 2023.

\bibitem{EVSSM}
Lingshun Kong, Jiangxin Dong, Jinhui Tang, Ming-Hsuan Yang, and Jinshan Pan.
\newblock Efficient visual state space model for image deblurring.
\newblock In {\em CVPR}, 2025.

\bibitem{10.1145/3065386}
Alex Krizhevsky, Ilya Sutskever, and Geoffrey~E. Hinton.
\newblock Imagenet classification with deep convolutional neural networks.
\newblock {\em Commun. ACM}, 60(6):84–90, may 2017.

\bibitem{labs2025flux1kontextflowmatching}
Black~Forest Labs, Stephen Batifol, Andreas Blattmann, Frederic Boesel, Saksham Consul, Cyril Diagne, Tim Dockhorn, Jack English, Zion English, Patrick Esser, Sumith Kulal, Kyle Lacey, Yam Levi, Cheng Li, Dominik Lorenz, Jonas Müller, Dustin Podell, Robin Rombach, Harry Saini, Axel Sauer, and Luke Smith.
\newblock Flux.1 kontext: Flow matching for in-context image generation and editing in latent space, 2025.

\bibitem{aim2025rawdenoising}
Feiran Li, Jiacheng Li, Marcos Conde, Beril Besbinar, Vlad Hosu, Daisuke Iso, Radu Timofte, et~al.
\newblock Real-world raw denoising using diverse cameras: {AIM} 2025 challenge report.
\newblock In {\em Proceedings of the IEEE/CVF International Conference on Computer Vision (ICCV) Workshops}, 2025.

\bibitem{liu2021swin}
Ze Liu, Yutong Lin, Yue Cao, Han Hu, Yixuan Wei, Zheng Zhang, Stephen Lin, and Baining Guo.
\newblock Swin transformer: Hierarchical vision transformer using shifted windows.
\newblock In {\em ICCV}, 2021.

\bibitem{aim2025perceptual}
Bruno Longarela, Marcos Conde, Álvaro García, Radu Timofte, et~al.
\newblock {AIM} 2025 perceptual image super-resolution challenge.
\newblock In {\em Proceedings of the IEEE/CVF International Conference on Computer Vision (ICCV) Workshops}, 2025.

\bibitem{cosine}
Ilya Loshchilov and Frank Hutter.
\newblock {SGDR:} stochastic gradient descent with warm restarts.
\newblock In {\em {ICLR}}, 2017.

\bibitem{Lu_2025_CVPR_EvenFormer}
Xin Lu, Yuanfei Bao, Jiarong Yang, Anya Hu, Jie Xiao, Kunyu Wang, Dong Li, Senyan Xu, Kean Liu, Xueyang Fu, and Zheng-Jun Zha.
\newblock Evenformer: Dynamic even transformer for real-world image restoration.
\newblock In {\em Proceedings of the Computer Vision and Pattern Recognition Conference (CVPR) Workshops}, pages 1081--1091, June 2025.

\bibitem{lu2025elucidating}
Xin Lu, Xueyang Fu, Jie Xiao, Zihao Fan, Yurui Zhu, and Zheng-Jun Zha.
\newblock Elucidating and endowing the diffusion training paradigm for general image restoration.
\newblock {\em arXiv preprint arXiv:2506.21722}, 2025.

\bibitem{Lu2025Tone}
Xin Lu, Yufeng Peng, Chengjie Ge, Zhijing Sun, Ziang Zhou, Zihao Li, Zishun Liao, Dong Li, Qiyu Kang, Xueyang Fu, and Zheng-Jun Zha.
\newblock Boosting inverse tone mapping via diffusion regularization.
\newblock In {\em 2025 {IEEE/CVF} International Conference on Computer Vision (ICCV) Workshops}. {IEEE} Computer Society, 2025.

\bibitem{Lu2025Deblurring}
Xin Lu, Zhijing Sun, Chengjie Ge, Yufeng Peng, Ziang Zhou, Zihao Li, Zishun Liao, Dong Li, Qiyu Kang, Xueyang Fu, and Zheng-Jun Zha.
\newblock Efficient high fps non-uniform motion deblurring via progressive learning.
\newblock In {\em 2025 {IEEE/CVF} International Conference on Computer Vision (ICCV) Workshops}. {IEEE} Computer Society, 2025.

\bibitem{Lu_2025_CVPR_ILAWR}
Xin Lu, Jie Xiao, Yurui Zhu, and Xueyang Fu.
\newblock Continuous adverse weather removal via degradation-aware distillation.
\newblock In {\em Proceedings of the Computer Vision and Pattern Recognition Conference (CVPR)}, pages 28113--28123, June 2025.

\bibitem{Lu_2025_CVPR_AALN}
Xin Lu, Jiarong Yang, Yuanfei Bao, Zihao Fan, Anya Hu, Kunyu Wang, Jie Xiao, Xi Wang, Hongjian Liu, Xueyang Fu, and Zheng-Jun Zha.
\newblock Advancing ambient lighting normalization via diffusion shadow generation.
\newblock In {\em Proceedings of the Computer Vision and Pattern Recognition Conference (CVPR) Workshops}, pages 1070--1080, June 2025.

\bibitem{Lu_2024_CVPR_HirFormer}
Xin Lu, Yurui Zhu, Xi Wang, Dong Li, Jie Xiao, Yunpeng Zhang, Xueyang Fu, and Zheng-Jun Zha.
\newblock Hirformer: Dynamic high resolution transformer for large-scale image shadow removal.
\newblock In {\em Proceedings of the IEEE/CVF Conference on Computer Vision and Pattern Recognition (CVPR) Workshops}, pages 6513--6523, June 2024.

\bibitem{LoFormer}
Xintian Mao, Jiansheng Wang, Xingran Xie, Qingli Li, and Yan Wang.
\newblock Loformer: Local frequency transformer for image deblurring.
\newblock In Jianfei Cai, Mohan~S. Kankanhalli, Balakrishnan Prabhakaran, Susanne Boll, Ramanathan Subramanian, Liang Zheng, Vivek~K. Singh, Pablo C{\'{e}}sar, Lexing Xie, and Dong Xu, editors, {\em Proceedings of the 32nd {ACM} International Conference on Multimedia, {MM} 2024, Melbourne, VIC, Australia, 28 October 2024 - 1 November 2024}, pages 10382--10391. {ACM}, 2024.

\bibitem{nah2017deep}
Seungjun Nah, Tae Hyun~Kim, and Kyoung Mu~Lee.
\newblock Deep multi-scale convolutional neural network for dynamic scene deblurring.
\newblock In {\em Proceedings of the IEEE conference on computer vision and pattern recognition}, pages 3883--3891, 2017.

\bibitem{GoPro}
Seungjun Nah, Tae~Hyun Kim, and Kyoung~Mu Lee.
\newblock Deep multi-scale convolutional neural network for dynamic scene deblurring.
\newblock In {\em 2017 {IEEE} Conference on Computer Vision and Pattern Recognition, {CVPR} 2017, Honolulu, HI, USA, July 21-26, 2017}, pages 257--265. {IEEE} Computer Society, 2017.

\bibitem{RealBlur}
Jaesung Rim, Haeyun Lee, Jucheol Won, and Sunghyun Cho.
\newblock Real-world blur dataset for learning and benchmarking deblurring algorithms.
\newblock In Andrea Vedaldi, Horst Bischof, Thomas Brox, and Jan{-}Michael Frahm, editors, {\em Computer Vision - {ECCV} 2020 - 16th European Conference, Glasgow, UK, August 23-28, 2020, Proceedings, Part {XXV}}, volume 12370 of {\em Lecture Notes in Computer Science}, pages 184--201. Springer, 2020.

\bibitem{aim2025scvqa}
Nickolay Safonov, Mikhail Rakhmanov, Dmitriy Vatolin, Radu Timofte, et~al.
\newblock {AIM} 2025 challenge on screen-content video quality assessment: Methods and results.
\newblock In {\em Proceedings of the IEEE/CVF International Conference on Computer Vision (ICCV) Workshops}, 2025.

\bibitem{chronos}
Kron Technologies.
\newblock Chronos 2.1-hd high-speed camera.
\newblock \url{https://www.krontech.ca/product/chronos-2-1-hd-high-speed-camera/}, 2024.
\newblock Accessed: 2024-11-11.

\bibitem{aim2025tone}
Chao Wang, Francesco Banterle, Bin Ren, Radu Timofte, et~al.
\newblock {AIM} 2025 challenge on inverse tone mapping report: Methods and results.
\newblock In {\em Proceedings of the IEEE/CVF International Conference on Computer Vision (ICCV) Workshops}, 2025.

\bibitem{wang2004image}
Zhou Wang, Alan~C Bovik, Hamid~R Sheikh, and Eero~P Simoncelli.
\newblock Image quality assessment: from error visibility to structural similarity.
\newblock {\em IEEE transactions on image processing}, 13(4):600--612, 2004.

\bibitem{Codabench}
Zhen Xu, Sergio Escalera, Adrien Pavão, Magali Richard, Wei-Wei Tu, Quanming Yao, Huan Zhao, and Isabelle Guyon.
\newblock Codabench: Flexible, easy-to-use, and reproducible meta-benchmark platform.
\newblock {\em Patterns}, 3(7):100543, 2022.

\bibitem{aim2025videodenoising}
Alexander Yakovenko, George Chakvetadze, Ilya Khrapov, Maksim Zhelezov, Dmitry Vatolin, Radu Timofte, et~al.
\newblock {AIM} 2025 low-light raw video denoising challenge: Dataset, methods and results.
\newblock In {\em Proceedings of the IEEE/CVF International Conference on Computer Vision (ICCV) Workshops}, 2025.

\bibitem{ESDNet}
Xin Yu, Peng Dai, Wenbo Li, Lan Ma, Jiajun Shen, Jia Li, and Xiaojuan Qi.
\newblock Towards efficient and scale-robust ultra-high-definition image demoir{\'{e}}ing.
\newblock In Shai Avidan, Gabriel~J. Brostow, Moustapha Ciss{\'{e}}, Giovanni~Maria Farinella, and Tal Hassner, editors, {\em Computer Vision - {ECCV} 2022 - 17th European Conference, Tel Aviv, Israel, October 23-27, 2022, Proceedings, Part {XVIII}}, volume 13678 of {\em Lecture Notes in Computer Science}, pages 646--662. Springer, 2022.

\bibitem{Restormer}
Syed~Waqas Zamir, Aditya Arora, Salman Khan, Munawar Hayat, Fahad~Shahbaz Khan, and Ming{-}Hsuan Yang.
\newblock Restormer: Efficient transformer for high-resolution image restoration.
\newblock In {\em CVPR}, 2022.

\bibitem{zamir2022restormer}
Syed~Waqas Zamir, Aditya Arora, Salman Khan, Munawar Hayat, Fahad~Shahbaz Khan, and Ming-Hsuan Yang.
\newblock Restormer: Efficient transformer for high-resolution image restoration.
\newblock In {\em Proceedings of the IEEE/CVF Conference on Computer Vision and Pattern Recognition (CVPR)}, pages 5728--5739, 2022.

\bibitem{zhang2018perceptual}
Richard Zhang, Phillip Isola, Alexei~A Efros, Eli Shechtman, and Oliver Wang.
\newblock The unreasonable effectiveness of deep features as a perceptual metric.
\newblock In {\em CVPR}, 2018.

\bibitem{zhang2025easycontrol}
Yuxuan Zhang, Yirui Yuan, Yiren Song, Haofan Wang, and Jiaming Liu.
\newblock Easycontrol: Adding efficient and flexible control for diffusion transformer.
\newblock {\em arXiv preprint arXiv:2503.07027}, 2025.

\end{thebibliography}
}

\clearpage
\appendix

\section{Teams and Affiliations}
\label{sec:teams}

\subsection*{AIM 2025 Team}
\noindent\textit{\textbf{Title: }} AIM 2025 High FPS Motion Deblurring Challenge\\
\noindent\textit{\textbf{Members: }} \\
George Ciubotariu$^1$
(\href{mailto:george.ciubotariu@uni-wuerzburg.de}{george.ciubotariu@uni-wuerzburg.de}),\\
Florin Vasluianu$^1$
(\href{mailto:florin-alexandru.vasluianu@uni-wuerzburg.de}{florin-alexandru.vasluianu@uni-wuerzburg.de}),\\
Zhuyun Zhou$^1$
(\href{mailto:zhuyun.zhou@uni-wuerzburg.de}{zhuyun.zhou@uni-wuerzburg.de}),\\
Nancy Mehta$^1$
(\href{mailto:nancy.mehta@uni-wuerzburg.de}{nancy.mehta@uni-wuerzburg.de}),\\
Radu Timofte$^1$ (\href{mailto:radu.timofte@uni-wuerzburg.de}{radu.timofte@uni-wuerzburg.de})\\
\noindent\textit{\textbf{Affiliations:}}\\
$^1$~Computer Vision Lab, CAIDAS \& IFI, Julius-Maximilians-Universit\"at of W\"urzburg, Germany\\

\subsection*{VPEG}
\noindent\textit{\textbf{Title: }} Enhanced Visual State Space Model for High FPS Non-Uniform Motion Deblur \\
\noindent\textit{\textbf{Members: }} \\
Ke Wu$^1$ 
(\href{mailto:wuke2002@njust.edu.cn}{wuke2002@njust.edu.cn}),\\
Long Sun$^1$ 
(\href{mailto:cs.longsun@njust.edu.cn}{cs.longsun@njust.edu.cn}),\\
Lingshun Kong$^1$ 
(\href{mailto:konglingshun@njust.edu.cn}{konglingshun@njust.edu.cn}),\\
Zhongbao Yang$^1$ 
(\href{mailto:yangzhongbao40@gmail.com}{yangzhongbao40@gmail.com}),\\
Jinshan Pan$^1$ 
(\href{mailto:jspan@njust.edu.cn}{jspan@njust.edu.cn}),\\
Jiangxin Dong$^1$ 
(\href{mailto:jxdong@njust.edu.cn}{jxdong@njust.edu.cn}),\\
Jinhui Tang$^1$ 
(\href{mailto:jinhuitang@njust.edu.cn}{jinhuitang@njust.edu.cn}),\\
\noindent\textit{\textbf{Affiliations: }} \\ 
$^1$~IMAG Lab, Nanjing University of Science and Technology\\

\subsection*{VPEG\_2\_3}
\noindent\textit{\textbf{Title: }} Efficient Frequency Domain-based Transformer and Discriminative FFN for High-Quality Image Deblurring (FFTformer+) \\
\noindent\textit{\textbf{Members: }} \\
Zhongbao Yang$^1$ 
(\href{mailto:yangzhongbao40@gmail.com}{yangzhongbao40@gmail.com}),\\
Hao Chen$^1$ 
(\href{mailto:jxpyy123@gmail.com}{jxpyy123@gmail.com}),\\
Yinghui Fang$^1$ 
(\href{mailto:floriayeahsmith@gmail.com}{floriayeahsmith@gmail.com}),\\
Jinshan Pan$^1$ 
(\href{mailto:jspan@njust.edu.cn}{jspan@njust.edu.cn}),\\
Jiangxin Dong$^1$ 
(\href{mailto:jxdong@njust.edu.cn}{jxdong@njust.edu.cn}),\\
Jinhui Tang$^1$ 
(\href{mailto:jinhuitang@njust.edu.cn}{jinhuitang@njust.edu.cn})\\
\noindent\textit{\textbf{Affiliations: }} \\ 
$^1$~IMAG Lab, Nanjing University of Science and Technology\\

\subsection*{SRC-B}
\noindent\textit{\textbf{Title: }} High FPS Motion Deblurring with Visual Prompt \\
\noindent\textit{\textbf{Members: }} \\
Dafeng Zhang$^1$ 
(\href{mailto:dfeng.zhang@samsung.com}{dfeng.zhang@samsung.com})\\
Yongqi Song$^1$
(\href{mailto:yongqi.song@samsung.com}{yongqi.song@samsung.com})\\
Jiangbo Guo$^1$
(\href{mailto:jiangbo.guo@samsung.com}{jiangbo.guo@samsung.com})\\
Shuhua Jin$^1$
(\href{mailto:shuhua.jin@samsung.com}{shuhua.jin@samsung.com})\\
\noindent\textit{\textbf{Affiliations: }} \\ 
$^1$Samsung R\&D Institute China-Beijing (SRC-B)\\
\subsection*{X\_L}
\noindent\textit{\textbf{Title: }} Multi-Scale Interaction Enhanced State Space Model for Image Deblurring \\
\noindent\textit{\textbf{Members: }} \\
Zeyu Xiao$^1$ 
(\href{mailto:zeyuxiao1997@163.com}{zeyuxiao1997@163.com }),\\
Rui Zhao$^2$ 
(\href{mailto:ruizhao26@gmail.com}{ruizhao26@gmail.com}),\\
Zhuoyuan Li$^3$ 
(\href{mailto:zhuoyuanli@mail.ustc.edu.cn}{zhuoyuanli@mail.ustc.edu.cn}),\\
Cong Zhang$^4$ 
(\href{mailto:cong-clarence.zhang@connect.polyu.hk}{cong-clarence.zhang@connect.polyu.hk})\\
\noindent\textit{\textbf{Affiliations: }} \\ 
$^1$National University of Singapore\\
$^2$Nanyang Technology University\\
$^3$University of Science and Technology of China\\
$^4$The Chinese University of Hong Kong\\

\subsection*{BlurKing\_Sharper}
\noindent\textit{\textbf{Title: }} Efficient High FPS Non-Uniform Motion Deblurring via Progressive Learning \\
\noindent\textit{\textbf{Members: }} \\
Yufeng Peng$^{1}$ (\href{mailto:pengyufeng@mail.ustc.edu.cn}{pengyufeng@mail.ustc.edu.cn})\\
Xin Lu$^{1}$ (\href{mailto:luxion@mail.ustc.edu.cn}{luxion@mail.ustc.edu.cn})\\
Zhijing Sun$^{1}$ (\href{mailto:sunzhijing@mail.ustc.edu.cn}{sunzhijing@mail.ustc.edu.cn})\\
Chengjie Ge$^{1}$ (\href{mailto:gechengjie@mail.ustc.edu.cn}{gechengjie@mail.ustc.edu.cn})\\
Zihao Li$^{1}$ (\href{mailto:lizihao@mail.ustc.edu.cn}{lizihao@mail.ustc.edu.cn})\\
Zishun Liao$^{1}$ (\href{mailto:liaozishun@mail.ustc.edu.cn}{liaozishun@mail.ustc.edu.cn})\\
Ziang Zhou$^{1}$ (\href{mailto:zhouziang@mail.ustc.edu.cn}{zhouziang@mail.ustc.edu.cn})\\
Qiyu Kang$^{1}$ (\href{mailto:kangqiyu@mail.ustc.edu.cn}{kangqiyu@mail.ustc.edu.cn})\\
 Xueyang Fu$^{1}$ (\href{mailto:fuxueyang@mail.ustc.edu.cn}{fuxueyang@mail.ustc.edu.cn})\\
 Zheng-Jun Zha$^{1}$ (\href{mailto:zhazhengjun@mail.ustc.edu.cn}{zhazhengjun@mail.ustc.edu.cn})\\
\noindent\textit{\textbf{Affiliations: }} \\
$^{1}$ School of Information Science and Technology and MoE Key Laboratory of Brain-inspired Intelligent Perception and Cognition, University of Science and Technology of China, Hefei, 230026, China\\ 

\subsection*{Mier}
\noindent\textit{\textbf{Title: }} EasyControl-Lite \\
\noindent\textit{\textbf{Members: }} \\
Yuqian Zhang (\href{mailto:zhangyuqian@xiaomi.com}{zhangyuqian@xiaomi.com})\\
Shuai Liu
(\href{mailto:liushuai@xiaomi.com}{liushuai@xiaomi.com})\\ 
\noindent\textit{\textbf{Affiliations: }} Xiaomi\\ 

\subsection*{BLR}
\noindent\textit{\textbf{Title: }} Image Deblurring Based on Pre-trained Model Fusion and Fine-tuning \\
\noindent\textit{\textbf{Members: }} \\
Jie Liu$^{1}$
(\href{mailto:liujie@nju.edu.cn}{liujie@nju.edu.cn})\\
Zhuhao Zhang$^{1}$
(\href{mailto:zhuhaozhang@smail.nju.edu.cn}{zhuhaozhang@smail.nju.edu.cn})\\
\noindent\textit{\textbf{Affiliations: }} \\
$^{1}$ Nanjing University, School of Computer Science\\ 

\subsection*{Nankai-CVLab}
\noindent\textit{\textbf{Title: }} RecuDiff-UT: Recursive Diffusion-Transformer with Self-Corrective Training\\
\noindent\textit{\textbf{Members: }} \\
Lishen Qu$^1$
(\href{mailto:qulishen@mail.nankai.edu.cn}{qulishen@mail.nankai.edu.cn}),\\
Zhihao Liu$^1$
(\href{mailto:2212602@mail.nankai.edu.cn}{2212602@mail.nankai.edu.cn}),\\
Shihao Zhou$^1$
(\href{mailto:zhoushihao96@mail.nankai.edu.cn}{zhoushihao96@mail.nankai.edu.cn}),\\
Yaqi Luo$^1$
(\href{mailto:570470068c@gmail.com}{570470068c@gmail.com}),\\
Juncheng Zhou$^1$ (\href{mailto:2112612@mail.nankai.edu.cn}{2112612@mail.nankai.edu.cn}),\\
Jufeng Yang$^1$
(\href{mailto:yangjufeng@nankai.edu.cn}{yangjufeng@nankai.edu.cn})\\
\noindent\textit{\textbf{Affiliations:}}\\
$^1$~Computer Vision Lab, Nankai University\\

\subsection*{MagicBlur}
\noindent\textit{\textbf{Title: }} Image Deblurring via Fine-Tuning with Pseudo Ground Truth Image\\
\noindent\textit{\textbf{Members: }} \\
Qianfeng Yang$^1$
(\href{mailto:csqianfengyang@163.com}{csqianfengyang@163.com}),\\
Qiyuan Guan$^1$
(\href{mailto:csguanqiyuan@163.com}{csguanqiyuan@163.com}),\\
Xiang Chen$^2$
(\href{mailto:chenxiang@njust.edu.cn}{chenxiang@njust.edu.cn}),\\
Guiyue Jin$^1$
(\href{mailto:guiyue.jin@dlpu.edu.cn}{guiyue.jin@dlpu.edu.cn}),\\
Jiyu Jin$^1$ (\href{mailto:jiyu.jin@dlpu.edu.cn}{jiyu.jin@dlpu.edu.cn})\\
\noindent\textit{\textbf{Affiliations:}}\\
$^1$~Dalian Polytechnic University\\
$^2$~Nanjing University of Science and Technology\\

\end{document}